\documentclass[10pt,conference]{IEEEtran}
\usepackage{amsmath,amssymb,amsfonts}
\usepackage{graphicx}
\usepackage{textcomp}
\usepackage{xcolor}
\usepackage{cite}
\usepackage{booktabs}
\usepackage{multirow}
\usepackage{array}
\usepackage{url}

\begin{document}

\title{Enhancing Sentiment Classification with\\ Machine Learning and Combinatorial Fusion}

\author{\IEEEauthorblockN{Sean Patten\textsuperscript{†}, Pin-Yu Chen\textsuperscript{‡}, Christina Schweikert\textsuperscript{¶}, D. Frank Hsu\textsuperscript{†}}
\IEEEauthorblockA{\textsuperscript{†}Laboratory of Informatics and Data Mining, Department of Computer and Information Science,\\
Fordham University, New York, NY 10023, USA\\
\textsuperscript{‡}IBM Thomas J. Watson Research Center, Yorktown Heights, NY 10598, USA\\\textsuperscript{¶}Division of Computer Science, Mathematics and Science, St. John's University, Queens, NY 11439, USA\\
E-mail: spatten2@fordham.edu; pin-yu.chen@ibm.com; schweikc@stjohns.edu; hsu@fordham.edu}}

\maketitle

\begin{abstract}
This paper presents a novel approach to sentiment classification using the application of Combinatorial Fusion Analysis (CFA) to integrate an ensemble of diverse machine learning models, achieving state-of-the-art accuracy on the IMDB sentiment analysis dataset of 97.072\%. CFA leverages the concept of cognitive diversity, which utilizes rank-score characteristic functions to quantify the dissimilarity between models and strategically combine their predictions. This is in contrast to the common process of scaling the size of individual models, and thus is comparatively efficient in computing resource use. Experimental results also indicate that CFA outperforms traditional ensemble methods by effectively computing and employing model diversity. The approach in this paper implements the combination of a transformer-based model of the RoBERTa architecture with traditional machine learning models, including Random Forest, SVM, and XGBoost. 
\end{abstract}

\section{Introduction}

Sentiment analysis, also commonly referred to as opinion mining, represents a significant area within Natural Language Processing (NLP). Its primary goal is to computationally interpret and classify the emotions, opinions, and attitudes expressed within textual data \cite{liu2012sentiment,pang2008opinion}. At its core, sentiment analysis identifies whether the underlying sentiment conveyed in a piece of text is positive, negative, or neutral, often extending to finer-grained levels of intensity or specific emotions \cite{liu2012sentiment}.

Despite its utility, sentiment analysis faces considerable challenges. Human language is inherently complex and nuanced. Accurately capturing sentiment requires understanding context, domain-specific terminology, sarcasm, irony, and implicit expressions of opinion \cite{liu2012sentiment,pang2008opinion,alahmadi2025generalizing,islam2020review}.

Various computational methods have been developed to tackle sentiment analysis. Early approaches often relied on predefined lexicons or rule-based systems \cite{liu2012sentiment,pang2008opinion}. While useful, these methods can struggle with context-dependency and domain adaptation. Subsequently, traditional machine learning (ML) algorithms, such as Support Vector Machines (SVM) \cite{cortes1995support}, Naive Bayes, and Random Forest (RF) \cite{breiman2001random}, applied to text features (like TF-IDF), became prevalent \cite{liu2012sentiment,pang2008opinion}.

More recently, the field has been revolutionized by the advent of deep learning, particularly transformer-based architectures like BERT (Bidirectional Encoder Representations from Transformers) \cite{devlin2019bert}. BERT and its successors learn contextualized word representations by processing text bidirectionally, allowing them to capture subtle meanings dependent on surrounding words \cite{devlin2019bert,rogers2020primer}. RoBERTa (Robustly Optimized BERT Pretraining Approach) represents a significant refinement of BERT, achieved through modifications to the pretraining strategy \cite{liu2019roberta}.

While individual models, especially large pre-trained language representation models, can be powerful, another effective strategy for enhancing predictive performance, robustness, and generalization in machine learning is ensemble learning \cite{dietterich2000ensemble,zhou2012ensemble,etelis2024generating}. This technique operates on the principle that aggregating the predictions of multiple individual models (learners) can yield results superior to any single constituent model \cite{dietterich2000ensemble}. This strategy helps to mitigate the bias-variance dilemma, where a single complex model might overfit (high variance) and a simple one might underfit (high bias) \cite{geman1992neural}.

A fundamental principle underpinning ensemble learning is that combining diverse, or more precisely mathematically independent models, that make different errors or capture different aspects of the data, is crucial for improving overall performance \cite{dietterich2000ensemble,kuncheva2003measures}. However, traditional ensemble techniques may not optimally exploit this diversity, particularly when combining models with heterogeneous architectures or varying performance levels. The underlying idea of combining model outputs for improved performance, often called consensus scoring, has proven effective in many data-intensive scientific domains and of particular note is often used in virtual screening in cheminformatics \cite{yang2005consensus}. The principles of fusing diverse models with CFA specifically have been successfully applied in computationally intensive fields such as drug discovery, where it has been used to enhance the prediction of ADMET (absorption, distribution, metabolism, excretion, and toxicity) properties to achieve top results on standard benchmarks \cite{jiang2023enhancing}. 

Combinatorial Fusion Analysis (CFA) offers a sophisticated paradigm specifically designed for combining multiple scoring systems (MSS) which are essentially any models or processes that assign scores to data items \cite{hsu2006combinatorial}. Proposed initially by Hsu, Chung, and Kristal \cite{hsu2006combinatorial}, CFA provides a structured workflow that goes beyond simple aggregation methods. A core element of CFA is the characterization of each scoring system using its rank-score Characteristic (RSC) function \cite{hsu2010rank}. 

Building upon the RSC function, CFA introduces the concept of Cognitive Diversity (CD) \cite{hsu2010rank}. Developed further by Hsu, Kristal, and Schweikert, CD quantifies the dissimilarity between the scoring behaviors of two systems, as captured by their RSC functions \cite{hsu2010rank,hsu2019cognitive,Hurley2021}. CFA utilizes the CD measure to inform the selection of systems for fusion and to guide the combination process itself, offering algorithms for both score and rank combination \cite{hsu2006combinatorial}.

This paper presents the application of Combinatorial Fusion Analysis to the task of sentiment classification. Specifically, it details the fusion of a state-of-the-art transformer model, RoBERTa \cite{liu2019roberta}, with a set of diverse and powerful traditional machine learning models: Random Forest \cite{breiman2001random}, Support Vector Machine \cite{cortes1995support}, and XGBoost \cite{chen2016xgboost}. The primary contribution is demonstrating that this CFA-driven fusion achieves exceptionally high accuracy (97.072\%) on the standard IMDb movie review benchmark dataset \cite{maas2011learning}.

\section{Methods and Experiments}

\subsection{Base Models and Feature Representations}
Our ensemble is composed of four base models: RoBERTa \cite{liu2019roberta}, SVM \cite{cortes1995support}, XGBoost \cite{chen2016xgboost}, and Random Forest \cite{breiman2001random}. These models were chosen to provide a broad spectrum of algorithmic approaches, from deep neural networks to classic machine learning, thereby ensuring a high degree of diversity among the models. The model details are as follows:

\textbf{RoBERTa (Model A):} 
RoBERTa enhances BERT by using dynamic masking, where masked tokens change during training epochs, and removes the Next Sentence Prediction (NSP) task to focus solely on masked language modeling, resulting in more robust representations.
We utilized a pre-trained RoBERTa-base model fine-tuned on IMDb data (the HuggingFace model roberta-base-imdb) as our neural network classifier \cite{chang2021roberta}. This model achieved about 94.67\% accuracy on the IMDb test set on its own.

\textbf{SVM (Model B):} SVMs are effective in handling high-dimensional feature spaces, as is common in text classification. The use of a linear kernel is computationally efficient and performs well for this task, especially with bag-of-words features. We used an SVM with a linear kernel trained on bag-of-words features extracted from the IMDb reviews. In our experiments, the SVM model achieved 83.46\% accuracy on the IMDb test set. 

\textbf{XGBoost (Model C):} XGBoost is an optimized gradient boosting method that iteratively trains decision trees, with each new tree attempting to correct the errors made by the previous ones. This sequential process creates a strong predictive model. We used XGBoost with default parameters on the same bag-of-words features. XGBoost obtained 85.36\% accuracy on the IMDb test set.

\textbf{Random Forest (Model D):} Random Forest creates a diverse set of decision trees by using bootstrap aggregating (bagging) and random feature selection at each node. This process helps to decorrelate the trees and reduce overfitting. We trained a Random Forest classifier on bag-of-words features. The Random Forest achieved 86.15\% accuracy on the test set, outperforming the other classical models.

It is important to note that all three classical models (SVM, Random Forest, XGBoost) use the same input representation (bag-of-words counts), whereas RoBERTa uses a completely different representation (subword tokenization into an embedding space) \cite{liu2019roberta}. This difference in feature representation is a deliberate choice to maximize data diversity without including additional datasets.

\subsection{Understanding Model Architectural Diversity}

The effectiveness of our ensemble approach stems fundamentally from the architectural diversity of the constituent models. Each model type processes and interprets textual data through fundamentally different computational paradigms, creating complementary strengths and weaknesses that CFA can exploit.

RoBERTa operates through self-attention mechanisms that allow it to model complex, long-range dependencies in text. The transformer architecture enables the model to understand context bidirectionally, meaning that each word's representation is influenced by all other words in the sequence. This allows RoBERTa to capture nuanced semantic relationships, such as understanding that "not bad" often means "good" in informal contexts, or recognizing when sarcasm inverts the literal meaning of words.

In contrast, Random Forest approaches the problem through an ensemble of decision trees, each making binary splits based on the presence or frequency of specific words. While this may seem simplistic compared to neural approaches, Random Forest excels at identifying robust patterns in word co-occurrences. For instance, it might learn that reviews containing both "waste" and "time" together are strongly indicative of negative sentiment, regardless of the specific context or word order.

Support Vector Machines take yet another approach, seeking to find an optimal hyperplane in the high-dimensional bag-of-words space that maximally separates positive and negative reviews. The SVM's strength lies in its ability to handle the curse of dimensionality effectively - even with thousands of unique words creating a vast feature space, SVMs can find meaningful decision boundaries. The linear kernel we employ essentially learns a weighted combination of word frequencies, where the weights indicate each word's importance for sentiment classification.

XGBoost, as a gradient boosting method, builds its model iteratively, with each new tree specifically targeting the errors made by the previous ensemble. This creates a highly adaptive model that can capture non-linear patterns in the data. Unlike Random Forest's parallel tree construction, XGBoost's sequential approach allows it to focus computational resources on the most challenging cases, potentially capturing subtle patterns that other models miss.

This architectural diversity translates directly into diverse error patterns. When RoBERTa misclassifies a review, it's often because the sentiment is expressed through unusual linguistic constructions or requires world knowledge beyond its training. The classical models, operating on simpler word frequency patterns, might correctly classify these same reviews if they contain telltale vocabulary. Conversely, reviews with complex contextual sentiment that the classical models struggle with may be easily handled by RoBERTa's sophisticated language understanding.

\subsection{Combinatorial Fusion Analysis (CFA)}
The core methodology for integrating the predictions of the four base models is Combinatorial Fusion Analysis (CFA) \cite{hsu2006combinatorial}\cite{hsu2024combinatorial}. CFA provides a mathematical framework for characterizing and combining multiple scoring systems (MSS) using the following concepts and methods:

\textbf{Scoring Systems:} In the CFA framework, each base model is treated as a scoring system. For a given dataset $D=\{d_1,d_2,...,d_n\}$ containing $n$ data items (movie reviews), a scoring system $A$ is defined by two functions \cite{hsu2010rank}:
\begin{itemize}
\item A score function $s_A : D \rightarrow \mathbb{R}$, which assigns a real-valued score.
\item A rank function $r_A : D \rightarrow \mathbb{N}$, which assigns a rank in the set of natural numbers.
\end{itemize}

\textbf{Rank-Score Characteristic (RSC) Function:} The RSC function, $f_A$, characterizes the intrinsic scoring behavior of system $A$ by linking ranks to scores \cite{hsu2010rank}. It allows us to derive ranks from scores for any given dataset item $i$:
$$f_A(i) = s_A(r_A^{-1}(i)) = (s_A \circ r_A^{-1})(i), \quad i \in \mathbb{N}$$

\textbf{Cognitive Diversity (CD):} Cognitive Diversity, also known as rank-score diversity, provides a quantitative measure of the dissimilarity between the scoring behavior of two systems, $A$ and $B$, based on their respective RSC functions \cite{hsu2019cognitive}. A higher CD value indicates greater dissimilarity:
$$cd(A, B) = d(f_A, f_B) = \left(\frac{\sum_{i=1}^{n}(f_A(i) - f_B(i))^2}{n}\right)^{\frac{1}{2}}$$

\textbf{Diversity Strength (DS):} Diversity Strength for a scoring system is computed as the average cognitive diversity between that system and every other scoring system, or model, in the ensemble. This provides a measure that captures a model's overall diversity compared to the other models being considered:
$$ds(A_j) = \frac{\sum_{i=1,i\neq j}^{t} cd(A_j, A_i)}{t - 1}$$

\textbf{CFA Combination Methods:} CFA offers various algorithms to combine the scores or ranks from multiple systems \cite{hsu2006combinatorial}. It should be noted that normalization of prediction scores is required to calculate the below. This study explores the following methods \cite{hsu2006combinatorial,jiang2023enhancing}:

1. \textbf{Average Combination (AC):} Average score combination and rank combination are used. For each we combine either the score or rank functions with equal weight to get an aggregate score or rank function:
$$s_{SC}(d_i) = \frac{\sum_{j=1}^{t} s_{A_j}(d_i)}{t} \quad \text{and} \quad s_{RC}(d_i) = \frac{\sum_{j=1}^{t} r_{A_j}(d_i)}{t}$$

2. \textbf{Weighted Combination by Performance (WCP):} Performance-weighted score combination and rank combination are also used. We combine either the score or ranks functions, weighted by individual base model performance to obtain an aggregate score or rank function:

$$
s_{SC}(d_i)=\frac{\sum^t_{j=1}p(A_j)(s_{A_j}(d_i))}{\sum^t_{j=1}p(A_j)} s_{RC}$$
$$
(d_i)=\frac{\sum^t_{j=1}\frac{1}{p(A_j)}(r_{A_j}(d_i))}{\sum^t_{j=1}\frac{1}{p(A_j)}} s_{SC}$$

3. \textbf{Weighted Combination by Diversity Strength (WCDS):} We also generate score and rank combinations weighted by diversity strength. Here, we perform a weighted score or rank combination, using the measured diversity strength of the corresponding models as the weights, resulting in an aggregate score or rank function:

$$
(d_i)=\frac{\sum^t_{j=1}ds(A_j)(s_{A_j}(d_i))}{\sum^t_{j=1}ds(A_j)} s_{RC}
$$
$$
(d_i)=\frac{\sum^t_{j=1}\frac{1}{ds(A_j)}(r_{A_j}(d_i))}{\sum^t_{j=1}\frac{1}{ds(A_j)}}
$$

\subsection{Dataset, Evaluation Metrics, and Experiments}
The dataset analyzed in this study is the widely recognized IMDb Movie Reviews dataset \cite{maas2011learning}. Originally compiled and introduced by Maas et al. (2011) \cite{maas2011learning}, it has become a standard benchmark for binary sentiment classification tasks \cite{maas2011learning,paperswithcode}. The dataset comprises 50,000 movie reviews collected from the Internet Movie Database (IMDb) \cite{maas2011learning}. It is specifically designed for binary sentiment analysis, with reviews labeled as either positive or negative \cite{maas2011learning}. The dataset is balanced, containing exactly 25,000 positive and 25,000 negative reviews \cite{maas2011learning}. This work utilizes the standard 25k/25k train/test split.

The primary evaluation metric used is accuracy, reflecting the percentage of correctly classified reviews on the held-out test set. Recall, precision, and F1 score were also calculated.

The RoBERTa model was implemented using the Hugging Face Transformers library \cite{wolf2020transformers}. Random Forest and SVM were implemented using scikit-learn\cite{pedregosa2011scikit}, and XGBoost was implemented using its dedicated library. For RoBERTa \cite{liu2019roberta}, the standard tokenizer associated with the pre-trained model was used. For the traditional ML models, text reviews were converted into numerical features using bag-of-words vectorization. 

\subsection{Implementation Details and Hyperparameter Configuration}

The implementation of our ensemble required careful consideration of hyperparameters for each constituent model. Through systematic experimentation, we identified configurations that balance individual model performance with computational efficiency.

For the Random Forest model, we employed an unusually high number of estimators (25,000), matching the size of our test dataset. This configuration emerged from empirical observations showing that ensemble accuracy continued to improve with increasing estimators, even when individual Random Forest accuracy plateaued. This suggests that additional trees capture subtle patterns that, while not improving standalone accuracy, provide valuable diversity for ensemble fusion. The maximum tree depth was left unbounded, allowing trees to fully capture complex patterns in the data. Overfitting's impact was minimized in development by evaluating overfitting on the ensemble's performance without regard to that of individual models, recognizing that increasing the bias of individual models in an ensemble is acceptable if the ensemble properly corrects the aggregate result.

The SVM implementation utilized a linear kernel with the regularization parameter C set to 1.0. We enabled probability estimation, which is crucial for CFA as it requires confidence scores rather than just binary predictions. The linear kernel was chosen for its computational efficiency on high-dimensional text data and its proven effectiveness in text classification tasks.

XGBoost was configured with default parameters, including a learning rate of 0.3, maximum tree depth of 6, and 100 boosting rounds. The log loss objective function was used for binary classification. These default settings provided a good balance between model complexity and generalization ability.

RoBERTa is fine-tuned on IMDb, the SVM/XGBoost/Random Forest are trained on 20k bag-of-words features (linear SVM with probs, default XGBoost, RF up to 25k trees), and all outputs are min–max normalized on training ranges and ranked for CFA. This process reduces the impact of low frequency terms from affecting the decision making process. RoBERTa was chosen over BERT and DistilBERT base models due to superior model performance within local computation limits.

The feature extraction process for traditional models involved converting raw text to lowercase, tokenizing on word boundaries, and creating frequency vectors. We deliberately avoided more sophisticated preprocessing (such as stemming or TF-IDF weighting) to maintain clear differentiation from RoBERTa's subword tokenization approach, thereby maximizing feature-level diversity.

\section{Results and Discussion}

\subsection{Results}
As expected, RoBERTa (Model A) \cite{liu2019roberta} was by far the strongest single model, achieving 94.67\% accuracy with an F1-score of 0.9473. The classical models showed substantially lower accuracy. RandomForest (D) achieved 86.15\% accuracy, XGBoost (C) achieved 85.36\% accuracy, and SVM (B) had 83.46\% accuracy. The top individual models are shown in Table I below. Table II shows the performance of the best model combinations, the method of combination and the resulting accuracy, with significantly better performance than the individual models.

\begin{table}[htbp]
\centering
\caption{Individual Model Performance}
\begin{tabular}{lccccc}
\toprule
Model & Accuracy & Precision & Recall & F1 \\
\midrule
RoBERTa (A)       & 0.94668 & 0.936313 & 0.95856 & 0.947306 \\
SVM (B)           & 0.83464 & 0.843940 & 0.82112 & 0.832374 \\
XGBoost (C)       & 0.85360 & 0.842211 & 0.87024 & 0.855996 \\
RandomForest (D)  & 0.86148 & 0.854531 & 0.87128 & 0.862824 \\
\bottomrule
\end{tabular}
\end{table}

\begin{table}[htbp]
\centering
\caption{Top Performance of Combined Models}
\begin{tabular}{llc}
\toprule
Models & Method & Accuracy \\
\midrule
(A, B, D) & Diversity (Test) SC & \textbf{0.97072} \\
(A, C, D) & Diversity (Train) SC & 0.97012  \\
(A, C) & Diversity (Test) RC & 0.97008 \\
(A, C, D) & Diversity (Test) SC & 0.97008 \\
\bottomrule
\end{tabular}
\end{table}

By applying the fusion methods described earlier \cite{jiang2023enhancing}, we observed significant improvements over the baseline models. The CFA-based ensemble – specifically, the diversity-weighted score fusion (WCDS-SC) of RoBERTa, SVM, and RandomForest – achieved an accuracy of 97.072\% on the IMDb test set \cite{maas2011learning}. This is a substantial increase of about 2.4 percentage points over RoBERTa alone, which corresponds to reducing the error rate by almost half. It should also be noted that the best combination without relying on batch prediction achieves 97.012

While the paper reports the highest accuracy for the (A, B, D) trio, other combinations also performed exceptionally well. For instance, the combination of RoBERTa, XGBoost, and RandomForest (A, C, D) using diversity strength score weighting achieved 97.012\% accuracy, demonstrating that multiple high-performing combinations exist within the ensemble. This provides evidence that the results are not simply the result of a lucky combination but derive from a method that consistently makes combinations better. The fact that top results are all taken by diversity metric using combinations further provides evidence that the technique's application is a general improvement in combination results.

It is informative to compare different ensemble combinations and weightings. Notably, the best-performing combination did not require all four models: the highest accuracy was obtained by combining RoBERTa (A), SVM (B), and RandomForest (D) using diversity-strength score weighting (WCDS-SC). 

To assess CFA's effectiveness, we compared the diversity-strength weighting (WCDS) against a more conventional performance-based weighting (WCP) \cite{jiang2023enhancing}. Using the combination of all four models, performance-weighted score fusion achieved 94.90\% accuracy, which is notably less than the 97.072\% with diversity weighting on the best trio. These differences consistently favor the diversity-informed fusion, supporting the central premise of CFA: diversity matters \cite{hsu2010rank,hsu2019cognitive}.

Fig. 1, Fig. 2, and Fig. 3. visualize the performance of various model combinations and fusion methods, using accuracy as the metric.

\begin{figure*}[htbp]
\centering
\includegraphics[width=0.8\textwidth]{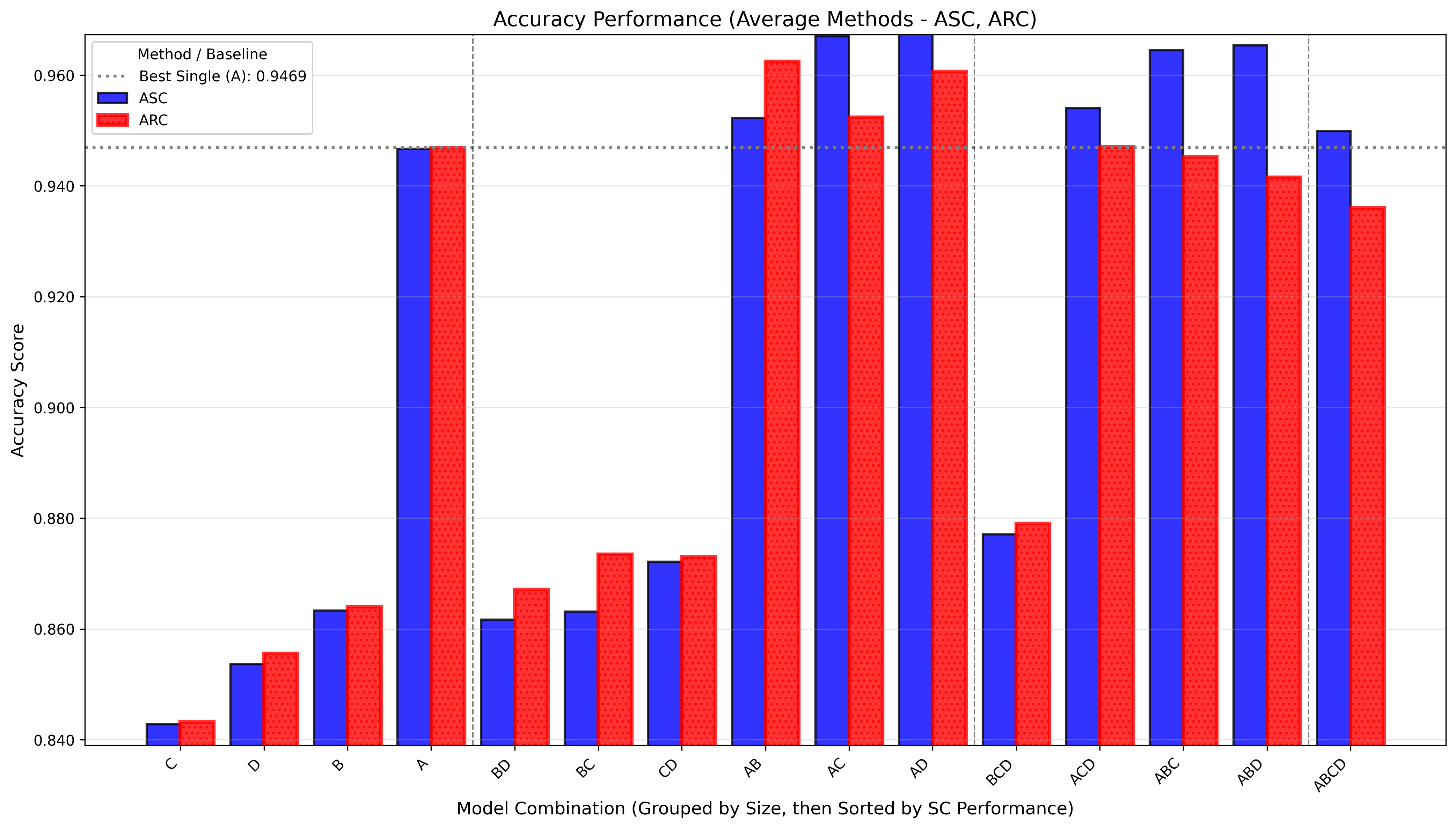}
\caption{Accuracy Performance (Average Methods - ASC, ARC).}
\label{fig:acc_avg}
\end{figure*}

\begin{figure*}[htbp]
\centering
\includegraphics[width=0.8\textwidth]{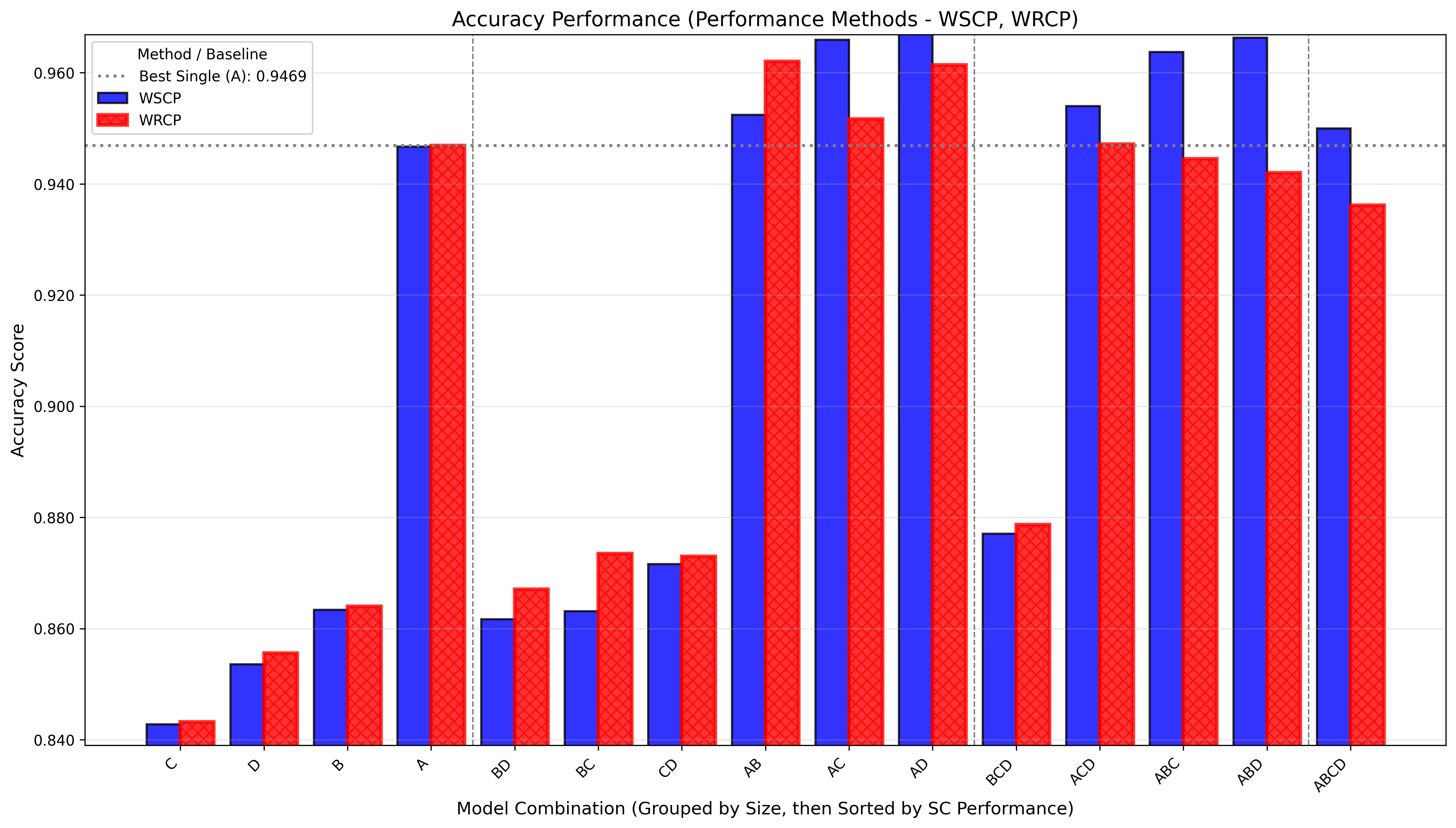}
\caption{Accuracy Performance (Performance Methods - WRCP, WRCDS).}
\label{fig:acc_perf}
\end{figure*}

\begin{figure*}[htbp]
\centering
\includegraphics[width=0.8\textwidth]{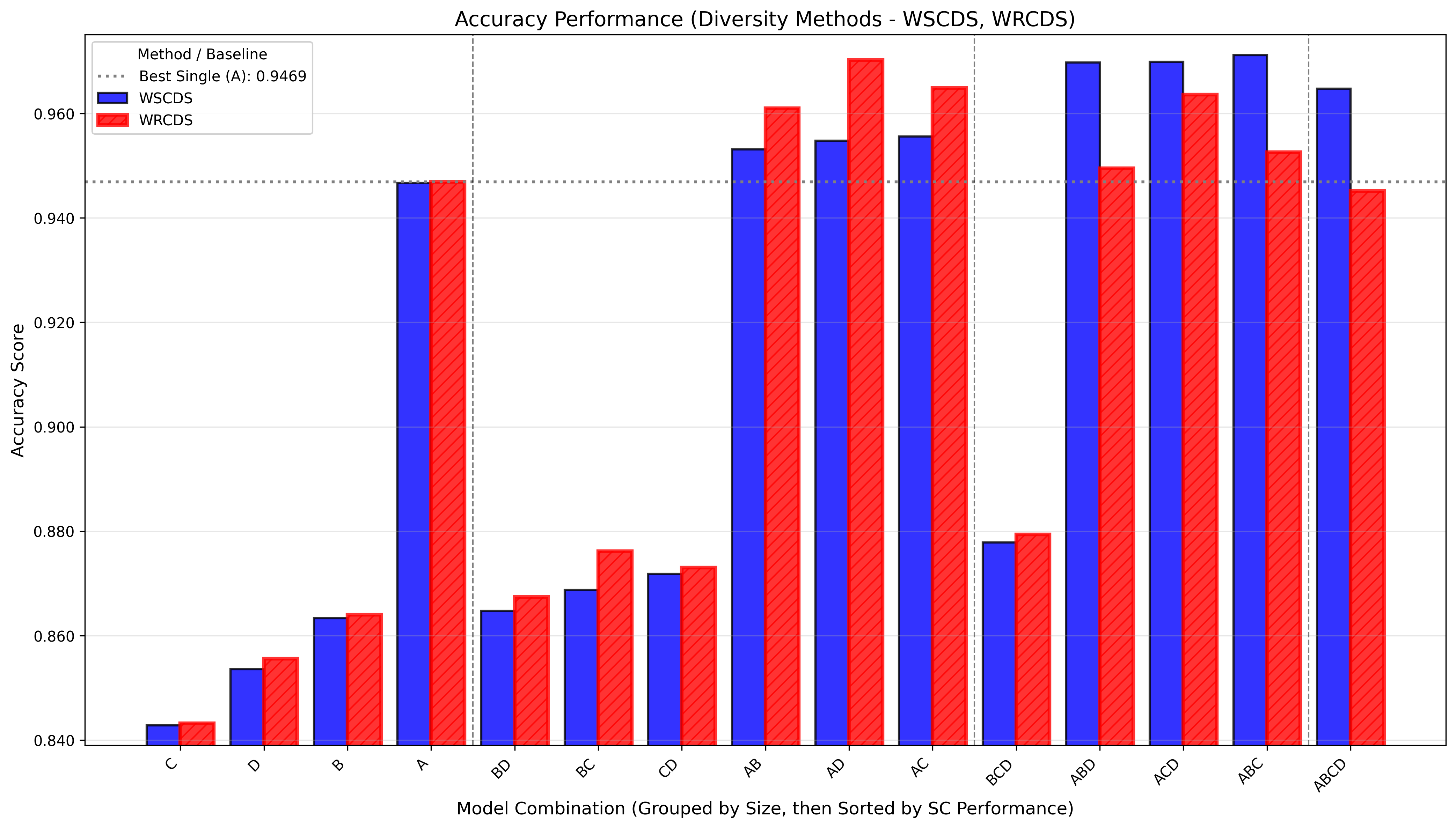}
\caption{Accuracy Performance (Diversity Methods - WSCDS, WRCDS).}
\label{fig:acc_div}
\end{figure*}

\subsection{Discussion}
Several insights emerged from our study that help explain its performance:

First, cognitive diversity was sufficiently high. The rank-score diversity analysis revealed significant differences in how the models rank the reviews which means that their combination is able to provide better results than less diverse models \cite{hsu2010rank,hsu2019cognitive,Hurley2021}. The cognitive diversity between RoBERTa and each classical model in particular was large – confirming that the transformer and bag-of-words approaches "view" the data differently. This can be clearly seen visually in the rank-score function graph below.

\begin{figure}[htbp]
\centering
\includegraphics[width=\columnwidth]{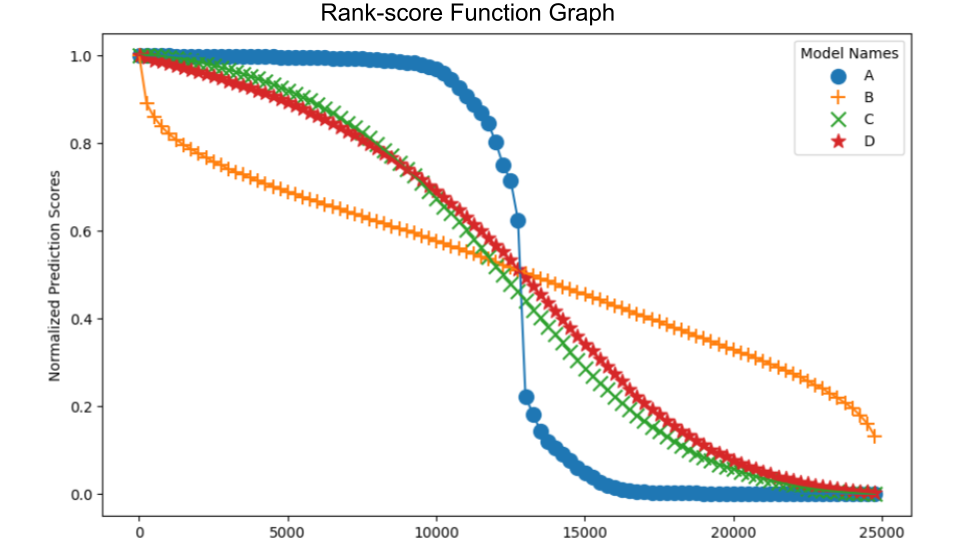}
\caption{Rank-score Function Graph by Model (Test Dataset). The graph shows the normalized prediction scores against the ranks for each of the four models: RoBERTa (A), SVM (B), XGBoost (C), and RandomForest (D).}
\label{fig:rank_score}
\end{figure}

Second, RoBERTa's presence was essential but not sufficient alone. Every top-performing combination included RoBERTa (A). If we removed RoBERTa and tried to combine just the three classical models (B,C,D), the accuracy maxed out around 87.8\%, far below RoBERTa's 94.67\%. However, adding any one of the other models to RoBERTa and fusing with diversity weights \cite{jiang2023enhancing} already gave a boost: (A,B), (A,C) or (A,D) each yielded between 95.5 - 97.0\% accuracy.

Third, ensemble component biases seem to have been resolved through model fusion. We examined the precision and recall of the ensemble compared to RoBERTa. RoBERTa had higher recall than precision (it favored predicting positive); the SVM had higher precision than recall (it was more conservative). In the fused result with the best combination (A, B, D), we ended up with precision 96.795\% and recall 97.368\%. This suggests the ensemble moderated RoBERTa's slight bias by incorporating models with complementary characteristics.

The rank-score function graph, Fig. 4, seen above provides a clear visualization of this diversity. RoBERTa (Model A) exhibits a steep sigmoidal curve, characteristic of a high-confidence classifier that separates positive and negative predictions with little difference in normalized score for most predictions. In contrast, the classical models (B, C, and D) show much flatter, more gradual RSC functions, indicating a less confident and more distributed scoring behavior. If these models were fundamentally making decisions in highly similar ways, we would not expect to see such different RSC functions. Recent work has explored various ensemble generation strategies for sentiment analysis \cite{etelis2024generating}, but our CFA-based approach differs in its calculation of diversity metrics. Specifically, as outlined in \cite{hsu2024combinatorial}, it ``is data, distribution, and domain independent since it is measured between two RSC functions, which are rank-centric." and the rank function carries more information than a score function on Euclidean space. It should also be noted that these are general to any implementation of the method and the steps used in this paper can generalize beyond the IMDb dataset and further work can be done to show performance on other benchmark sentiment datasets without any theoretical issues.

\subsection{Computational Efficiency and Practical Deployment}

An often overlooked aspect of ensemble methods is their computational footprint. Our approach demonstrates that significant performance gains can be achieved without proportional increases in computational costs. During inference, the CFA combination step itself is negligible, requiring simple arithmetic operations on the model outputs.

For practical deployment, the ensemble can be optimized further. The classical models can run on CPU while RoBERTa utilizes GPU resources, enabling efficient parallel processing. In scenarios where GPU access is limited, the bag-of-words features for classical models can be precomputed and cached, reducing repeated computation. At inference time, when models are executed in parallel, the end-to-end latency is effectively governed by the slowest model (typically RoBERTa), with the CFA fusion process adding negligible overhead, yielding throughput comparable to running the slowest model alone.

Due in part to the above, our result of 97.072\% accuracy slightly outperforms the previous reported best (96.68\% on IMDb) achieved by a RoBERTa-large model augmented with a large language model-based data annotation technique (the LlamBERT approach) \cite{csanady2024llambert}. The top ten models by classification accuracy posted on the IMDb Sentiment analysis leaderboard from paperswithcode.com as of June 2025 are listed in Table III for comparison. It should be noted that in the course of the publication of this paper that paperswithcode is no longer supported but the results can be verified by looking at individual papers introducing each model, links to which can be found on the internet archive of the page.

\begin{table}[htbp]
\scriptsize
\centering
\caption{Top 10 Models Accuracies on the IMDb Sentiment Analysis Leaderboard \cite{paperswithcode}}
\begin{tabular}{lcc}
\toprule
Model & Accuracy & Year \\
\midrule
RoBERTa-large with LlamBERT & 96.68 & 2024 \\
RoBERTa-large & 96.54 & 2024 \\
XLNet & 96.21 & 2019 \\
Heinsen Routing + RoBERTa Large & 96.2 & 2022 \\
RoBERTa-large 355M + Entailment as Few-shot Learner & 96.1 & 2021 \\
GraphStar & 96.0 & 2019 \\
DV-ngrams-cosine with NB sub-sampling + RoBERTa.base & 95.94 & 2022 \\
DV-ngrams-cosine + RoBERTa.base & 95.92 & 2022 \\
Roberta\_Large ST + Cosine Similarity Loss & 95.9 & 2025 \\
BERT large finetune UDA & 95.8 & 2019 \\
\bottomrule
\end{tabular}
\end{table}

An important practical aspect of model deployment is the computational cost. In our approach, we leverage a RoBERTa that was pre-trained and then fine-tuned on IMDb \cite{maas2011learning}. The other models (RF, SVM, XGBoost) are simpler classical ones. Thus the ensemble did not significantly add to the cost and time spent beyond RoBERTa's initial training and inference.

Our experiments were conducted on a consumer grade computer with a single NVIDIA RTX 4080 mobile GPU and an Intel i9 CPU, which is modest compared to the data center grade GPU servers often used to push NLP benchmarks \cite{strubell2019energy}. This highlights that CFA can offer a more accessible and energy-efficient route to state-of-the-art results fundamentally different from other more common performance scaling methods such as increasing training time.

\section{Conclusion}
We have presented a CFA-enhanced ensemble for sentiment analysis that achieves record-breaking performance on the IMDb movie review dataset. By combining a Transformer model (RoBERTa) \cite{liu2019roberta} with heterogeneous classifiers like SVM, XGBoost, and RandomForest through Combinatorial Fusion Analysis \cite{hsu2006combinatorial}, we attained an accuracy of 97.072\%, surpassing the accuracy of any individual model as well as the previous state-of-the-art \cite{csanady2024llambert}.

Our results demonstrate that carefully integrating diverse models can yield a whole that is greater than the sum of its parts \cite{dietterich2000ensemble}. The CFA ensemble corrected a wide array of mistakes that even a highly capable model like RoBERTa \cite{liu2019roberta} made, by amplifying the signals from other models that had complementary strengths \cite{kuncheva2003measures}. This led to substantial gains (over 2 percentage points in accuracy) that would be very difficult to achieve by other means short of enormous increases in training data or model size. The results also show that model fusion weighted by diversity strength of the models outperformed the more traditional approach using performance as a weight.

Beyond the specific accuracy numbers, this work highlights the importance of model diversity in NLP tasks \cite{kuncheva2003measures}. The success of our ensemble underscores that there is untapped information in models that might individually seem inferior. CFA provides a mechanism to unify these perspectives \cite{hsu2006combinatorial,hsu2010rank}, validating the notion that encouraging diversity in an ensemble can pay off significantly \cite{dietterich2000ensemble,kuncheva2003measures}.
This work also offers a practical and computationally efficient pathway to achieving state-of-the-art performance. The prevailing trend in NLP has been to scale model architectures to hundreds of millions or billions of parameters, a strategy that often yields diminishing returns and incurs substantial computational and environmental costs \cite{strubell2019energy}. Our approach demonstrates that by intelligently combining a moderately sized transformer with far less complex classical models, it is possible to surpass the performance of even larger, more resource-intensive systems. This makes high-performance sentiment analysis more accessible and sustainable, aligning with the growing need for efficient AI solutions.

It is important to acknowledge the limitations of this study, which in turn open avenues for future investigation. Our experiments were conducted exclusively on the IMDb dataset \cite{maas2011learning}, a benchmark known for its relatively long and well-structured reviews. The generalizability of our findings to other domains, such as short-form text from social media, product reviews, or financial news, remains to be validated. The linguistic characteristics of these domains differ significantly, and the optimal ensemble composition and fusion strategy may vary accordingly. Further research is needed to explore how cognitive diversity manifests and can be leveraged across different types of textual data.
Future research could extend this methodology in several promising directions. One natural step is to broaden the pool of base classifiers to include other diverse architectures, such as XLNet \cite{yang2019xlnet} or even non-neural, knowledge-based systems, to further probe the limits of cognitive diversity. Another promising direction is a deeper qualitative analysis of the errors corrected by the CFA fusion. By identifying the specific linguistic phenomena (e.g., sarcasm, complex negation, subtle contextual cues) that the ensemble successfully resolves, we can gain more profound insights into the complementary strengths of different model paradigms, moving beyond aggregate metrics to a more nuanced understanding of model behavior.
Ultimately, this study serves as a testament to the synergy between modern deep learning and classical machine learning paradigms when unified by a principled fusion framework. The success of this method reaffirms the foundational principles of ensemble learning: that a committee of diverse experts often outperforms any single genius \cite{dietterich2000ensemble}. It provides evidence to the claim that the intelligent fusion of cognitively diverse systems, as formalized by CFA \cite{hsu2006combinatorial}, represents not just a method for incremental improvement but a robust and sustainable strategy for advancing the capabilities of machine learning systems.

\section*{Acknowledgment}
The first author, Sean Patten, was supported by a Fordham-IBM Research Internship. 

\section*{Data and Code availability}
\url{https://github.com/seanpattencode/CFA4SA} is the source code used for this paper. Because Github restricts size of uploaded files, the cache of models is not included. The graphs are also not included. These will generate automatically if the code is run.

\bibliographystyle{IEEEtran}

\begin{thebibliography}{10}

\bibitem{alahmadi2025generalizing}
K.~Alahmadi, S.~Alharbi, J.~Chen, and X.~Wang, ``Generalizing sentiment analysis: A review of progress, challenges, and emerging directions,'' \emph{Social Network Analysis and Mining}, vol.~15, no.~1, p.~45, 2025.

\bibitem{breiman2001random}
L.~Breiman, ``Random forests,'' \emph{Machine Learning}, vol.~45, no.~1, pp. 5--32, 2001.

\bibitem{chang2021roberta}
A.~Chang, ``roberta-base-imdb,'' Hugging Face, 2021. [Online]. Available: \url{https://huggingface.co/textattack/roberta-base-imdb}

\bibitem{chen2016xgboost}
T.~Chen and C.~Guestrin, ``XGBoost: A scalable tree boosting system,'' in \emph{Proc. 22nd ACM SIGKDD Int. Conf. Knowl. Discovery Data Mining}, 2016, pp. 785--794.

\bibitem{cortes1995support}
C.~Cortes and V.~Vapnik, ``Support-vector networks,'' \emph{Machine Learning}, vol.~20, no.~3, pp. 273--297, 1995.

\bibitem{csanady2024llambert}
B.~Csanády \emph{et al.}, ``LlamBERT: Large-scale low-cost data annotation in NLP,'' \emph{arXiv preprint arXiv:2403.15938}, 2024.

\bibitem{devlin2019bert}
J.~Devlin, M.-W. Chang, K.~Lee, and K.~Toutanova, ``BERT: Pre-training of deep bidirectional transformers for language understanding,'' \emph{arXiv preprint arXiv:1810.04805}, 2018.

\bibitem{dietterich2000ensemble}
T.~G. Dietterich, ``Ensemble methods in machine learning,'' in \emph{Proc. Int. Workshop Multiple Classifier Systems}. Springer, 2000, pp. 1--15.

\bibitem{etelis2024generating}
I.~Etelis, A.~Rosenfeld, A.~I. Weinberg, and D.~Sarne, ``Generating effective ensembles for sentiment analysis,'' \emph{arXiv preprint arXiv:2402.16700}, 2024.

\bibitem{geman1992neural}
S.~Geman, E.~Bienenstock, and R.~Doursat, ``Neural networks and the bias/variance dilemma,'' \emph{Neural Computation}, vol.~4, no.~1, pp. 1--58, 1992.

\bibitem{hsu2006combinatorial}
D.~F. Hsu, S.~M. Chung, and B.~S. Kristal, ``Combinatorial fusion analysis: Methods and practices of combining multiple scoring systems,'' in \emph{Advanced Data Mining and Applications}. Springer, 2006, pp. 32--62.

\bibitem{hsu2010rank}
D.~F. Hsu, B.~S. Kristal, and C.~Schweikert, ``Rank-score characteristics (RSC) function and cognitive diversity,'' in \emph{Brain Informatics}. Springer, 2010, pp. 42--54.

\bibitem{hsu2019cognitive}
D.~F. Hsu, B.~S. Kristal, Y.~Hao, and C.~Schweikert, ``Cognitive diversity: A measurement of dissimilarity between multiple scoring systems,'' \emph{J. Interconnection Networks}, vol.~19, no.~04, p. 1940006, 2019.

\bibitem{hsu2024combinatorial}
D.~F. Hsu, B.~S. Kristal, and C.~Schweikert, ``Combinatorial fusion analysis,'' \emph{Computer}, vol.~57, no.~9, pp. 96--100, 2024.

\bibitem{Hurley2021}
L.~Hurley \emph{et al.}, ``Multi-layer combinatorial fusion using cognitive diversity,'' \emph{IEEE Access}, vol.~9, pp. 3919--3935, 2021.

\bibitem{islam2020review}
S.~Islam, N.~A. Ab Ghani, and M.~Ahmed, ``A review on recent advances in deep learning for sentiment analysis: Performances, challenges and limitations,'' \emph{Compusoft}, vol.~9, no.~7, pp. 3775--3783, 2020.

\bibitem{jiang2023enhancing}
N.~Jiang \emph{et al.}, ``Enhancing ADMET property models performance through combinatorial fusion analysis,'' \emph{J. Cheminformatics}, vol.~15, no.~1, pp. 1--18, 2023.

\bibitem{kuncheva2003measures}
L.~I. Kuncheva and C.~J. Whitaker, ``Measures of diversity in classifier ensembles,'' \emph{Machine Learning}, vol.~51, no.~2, pp. 181--207, 2003.

\bibitem{liu2012sentiment}
B.~Liu, ``Sentiment analysis and opinion mining,'' \emph{Synthesis Lectures on Human Language Technologies}, vol.~5, no.~1, pp. 1--167, 2012.

\bibitem{liu2019roberta}
Y.~Liu \emph{et al.}, ``RoBERTa: A robustly optimized BERT pretraining approach,'' \emph{arXiv preprint arXiv:1907.11692}, 2019.

\bibitem{maas2011learning}
A.~L. Maas \emph{et al.}, ``Learning word vectors for sentiment analysis,'' in \emph{Proc. 49th Annu. Meeting Assoc. Comput. Linguistics}, 2011, pp. 142--150.

\bibitem{pang2008opinion}
B.~Pang and L.~Lee, ``Opinion mining and sentiment analysis,'' \emph{Found. Trends Inf. Retr.}, vol.~2, no. 1-2, pp. 1--135, 2008.

\bibitem{pedregosa2011scikit}
F.~Pedregosa \emph{et al.}, ``Scikit-learn: Machine learning in Python,'' \emph{J. Machine Learning Research}, vol.~12, pp. 2825--2830, 2011.

\bibitem{paperswithcode}
Papers With Code, ``Sentiment Analysis on IMDb Leaderboard.'' [Online]. Available: \url{https://paperswithcode.com/sota/sentiment-analysis-on-imdb}

\bibitem{rogers2020primer}
A.~Rogers, O.~Kovaleva, and A.~Rumshisky, ``A primer in BERTology: What we know about how BERT works,'' \emph{arXiv preprint arXiv:2002.12327}, 2020.

\bibitem{strubell2019energy}
E.~Strubell, A.~Ganesh, and A.~McCallum, ``Energy and policy considerations for deep learning in NLP,'' in \emph{Proc. 57th Annu. Meeting Assoc. Comput. Linguistics}, 2019, pp. 3645--3650.

\bibitem{wolf2020transformers}
T.~Wolf \emph{et al.}, ``Transformers: State-of-the-art natural language processing,'' in \emph{Proc. Conf. Empirical Methods in Natural Lang. Process.: System Demonstrations}, 2020, pp. 38--45.

\bibitem{yang2005consensus}
J.-M. Yang \emph{et al.}, ``Consensus scoring criteria for improving enrichment in virtual screening,'' \emph{J. Chem. Inf. Model.}, vol.~45, no.~5, pp. 1134--1146, 2005.

\bibitem{yang2019xlnet}
Z.~Yang \emph{et al.}, ``XLNet: Generalized autoregressive pretraining for language understanding,'' in \emph{Adv. Neural Inf. Process. Syst.}, 2019, pp. 5753--5763.

\bibitem{zhou2012ensemble}
Z.-H. Zhou, \emph{Ensemble Methods: Foundations and Algorithms}. CRC Press, 2012.

\end{thebibliography}

\end{document}